\documentclass{article}

\usepackage[final]{neurips_2019}

\usepackage[utf8]{inputenc} 
\usepackage[T1]{fontenc}    
\usepackage{hyperref}       
\usepackage{url}            
\usepackage{booktabs}       
\usepackage{amsfonts}       
\usepackage{amsmath}
\usepackage{multirow}
\usepackage{multicol}
\usepackage{bm}
\usepackage{wrapfig}
\usepackage{todo}
\usepackage{nicefrac}       
\usepackage{microtype}      
\usepackage{caption}
\usepackage{algorithm}
\usepackage{mathabx}
\usepackage[noend]{algpseudocode}
\usepackage[T1]{fontenc}
\usepackage{verbatim}
\usepackage[utf8]{inputenc}
\usepackage[english]{babel}

\usepackage{microtype}
\usepackage{graphicx}
\usepackage{subfigure}
\usepackage{booktabs}
\usepackage[inline]{enumitem}

\title{Data Efficient Training for Reinforcement Learning with Adaptive Behavior Policy Sharing}

\author{%
  Ge Liu\thanks{work was done while an intern in Google} \\
  CSAIL, MIT \\
  \texttt{geliu@mit.edu} \\
  \And
  Rui Wu\\
  Google Brain\\
  \texttt{wrui@google.com} \\
  \AND
  Heng-Tze Cheng \\
  Google Brain\\
  \texttt{hengtze@google.com} \\
  \And
  Jing Wang \\
  Google Brain\\
  \texttt{jingconanwang@google.com} \\
  \And
  Jayden Ooi \\
  Google Brain\\
  \texttt{jayden@google.com} \\
  \And
  Lihong Li\\
  Google Brain\\
  \texttt{lihong@google.com} \\
  \And
  Ang Li \\
  DeepMind\\
  \texttt{anglili@google.com} \\
  \And
  Wai Lok Sibon Li \\
  DeepMind\\
  \texttt{sibon@google.com} \\
  \And
  Craig Boutilier\\
  Google AI\\
  \texttt{cboutilier@google.com} \\
  \And
  Ed Chi \\
  Google Brain\\
  \texttt{edchi@google.com} \\
}

\begin{document}

\newcommand{\methodname}{Adaptive Behavior Policy Sharing}
\newcommand{\methodabbrev}{ABPS}
\newcommand{\methodpbt}{ABPS-PBT}

\maketitle

\begin{abstract}
Deep Reinforcement Learning (RL) is proven powerful for decision making in simulated environments. However, training deep RL model is challenging in real world applications such as production-scale health-care or recommender systems because of the expensiveness of interaction and limitation of budget at deployment. One aspect of the data inefficiency comes from the expensive hyper-parameter tuning when optimizing deep neural networks. We propose Adaptive Behavior Policy Sharing (ABPS), a data-efficient training algorithm that allows sharing of experience collected by behavior policy that is adaptively selected from a pool of agents trained with an ensemble of hyper-parameters. We further extend ABPS to evolve hyper-parameters during training by hybridizing ABPS with an adapted version of Population Based Training (ABPS-PBT). We conduct experiments with multiple Atari games with up to $16$ hyper-parameter/architecture setups. ABPS achieves superior overall performance, reduced variance on top $25\%$ agents, and equivalent performance on the best agent compared to conventional hyper-parameter tuning with independent training, even though ABPS only requires the same number of environmental interactions as training a single agent. We also show that ABPS-PBT further improves the convergence speed and reduces the variance.
\end{abstract}

\section{Introduction}
\label{intro}
Recent years have witnessed the success of deep reinforcement learning (RL) in solving complex sequential decision making problems such as games\citep{mnih2013playing,SchulmanLMJA15,MnihBMGLHSK16}. However, it is yet proven to be effective in real world applications such as large-scale health-care or recommender system due to some practical constraints. First, there is no simulator available in many real-world problems and it is often expensive to do online interactions (e.g. through interacting with users). Second, only a small amount of online traffic and time budget is allocated for training new models. Last, the cost of deploying new models is too expensive for training to perform frequent behavior policy updates.

Hyper-parameter tuning and architecture design is key in the optimization of deep neural networks. However, unlike supervised learning, RL training requires subtle data-collection through interactions with the environment where traditional ways of hyper-parameter tuning with independent training shows their limits in terms of data efficiency. In this paper, we propose Adaptive Behavior Policy Sharing (ABPS), a data-efficient training algorithm for off-policy RL that allows sharing of experience collected by a single behavior policy that is adaptively selected from a pool of agents, which are trained with different hyper-parameters. 

Our main contributions are as follows:
1) We revisit the algorithm selection for RL with a new aspect: improved exploration due to the ensemble bootstrapping effect, and experiment with different exploration strategies for algorithm selection to study their effect on deep RL optimization.
2) We propose a more practical setup where the switching of behavior policy is less frequent.
3) We evaluate the best and the overall performance of the learned target policies instead of only evaluating the regret (behavior policy reward), because in real world only a single best agent is deployed of which we care about its expected reward. We show that the best agent converges as effective as the agent trained with independent training in terms of performance, and the pool of agents gain higher overall return with reduced variance on top $25\%$ agents.
4) We propose a novel adaptation of the Population Based Training (PBT) \citep{jaderberg2017population,Li2019} and integrate PBT with ABPS. This variant has the potential to efficiently train and evolve hyper-parameters simultaneously so that better hyper-parameter selection is achieved. We observe faster convergence with even lower variance and higher median on the top $25\%$ agents in the pool.

\section{Related Work} 
Many studies have focused on improving data-efficiency through better exploration using distributional value functions. \citet{bootstrap} introduced a multi-head Deep Q network (DQN) where each head is trained with bootstrapped data. They showed that with a randomly selected policy from the ensemble of DQN heads the exploration can be improved. Other recent studies in this direction introduced quantile regression~\citep{quantile} and distributional RL~\citep{BellemareDM17}. These works provide better uncertainty estimates for the value function, however, they do not address the hyper-parameter tuning problem.

Another strand of studies focus on utilizing the off-policy data. \citet{JiangL15,doubly2} used a doubly robust trick to improve off-policy evaluation of the state-action pairs. While this line of work has made the best use of off-policy data, there remains high bias and variance in the estimation inherited from the data.

Several studies have related algorithm selection~\citep{kotthoff2016algorithm} with reinforcement learning. \citet{Azar2013} showed that a bandit can choose the best RL policy with reasonable regret bound if a learned set of RL policies is given. However it does not taking the training and optimization process into account since the policies are fixed. \citet{laroche2018} proposed a similar framework for optimization algorithm selection in RL, and provided theoretical analysis on the regret bound of behavior policy. However the RL algorithm itself is considered as a black-box, and the paper has limited focus on how algorithm selection is affecting the exploration of RL and the performance of resulting target policies. In addition it adopted a setting which requires very frequent behavior policy selection.

\section{Methods}
\label{method}
\subsection{General problem setup}

We consider tasks in which an RL agent learns through sequential interactions with an environment $E$ whose internal state is unobservable. At each time-step $t\in \mathbb{N}$, the agent selects an action from a legal action set $a(t)\in \mathcal{A}$ according to policy $\pi(t)$, and then receives from $E$ a reward $r(t+1) \in \mathbb{R}$ and an observation $o(t+1)$ which is determined by the next state $s(t+1)$. 
\begin{wrapfigure}{r}{0.46\textwidth}
  \begin{center}
    \includegraphics[width=0.47\textwidth, trim=1cm 0.5cm 0cm 2cm]{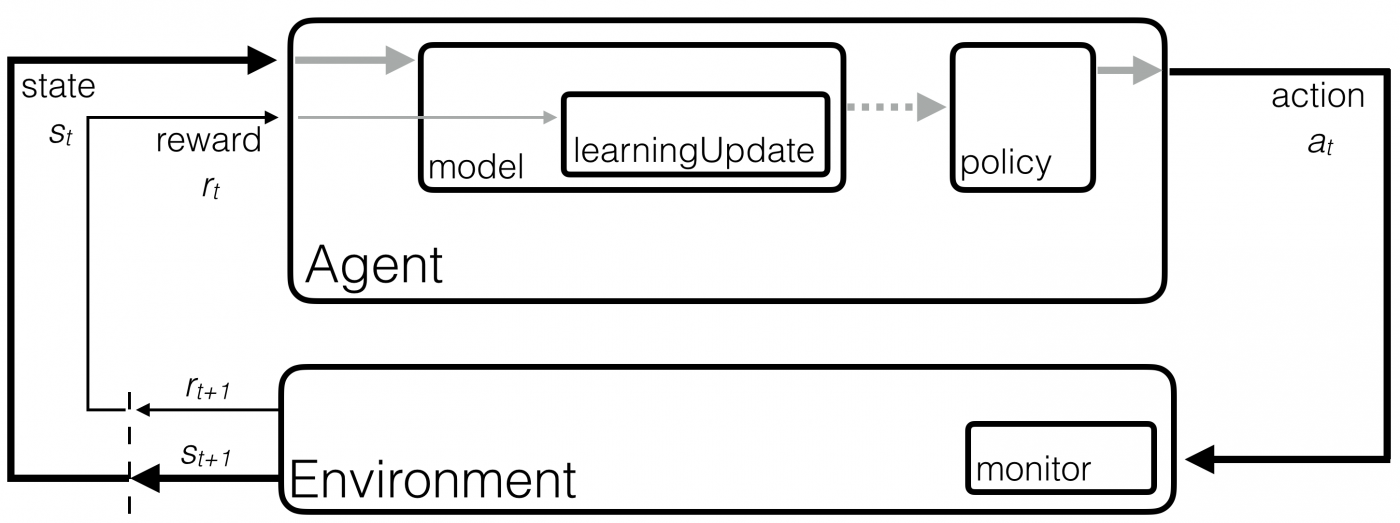}
  \end{center}
  \caption{RL framework}
\end{wrapfigure}
We assume the RL problem is episodic since it is a common set-up in games and real world problems, and policy selection on non-episodic RL is known to be hard \citep{Azar2013}. To ensure the algorithm selection satisfies the ``\textit{leaning is fair}'' assumption mentioned in \citep{laroche2018}, we mainly consider \textit{off-policy} RL algorithms, even though the sharing of behavior policy can be adapted to \textit{on-policy} algorithms with corrections ( \textit{e.g.}, importance sampling). For the purpose of demonstration, we further constrain our focus to Deep Q network (DQN)\cite{mnih2013playing}, which uses a neural network function approximator with weights $\theta$ as a Q-network to estimate the action-value function $Q(s,a;\theta)$, and utilizes a \textit{replay buffer} $\mathcal{D}$ to store the agent's experiences $e_t=<s_t,a_t,r_t,s_{t+1}>$ at each time-step. It learns a greedy policy $a=\arg\max_aQ(s,a;\theta)$, but an $\epsilon$-greedy policy of the $Q$ is often used during training with annealing $\epsilon$.

\begin{figure}[hbt!]
\vskip -0.1in
\centering
  \includegraphics[width=1\linewidth,trim=0cm 0cm 0cm 0cm, clip=true]{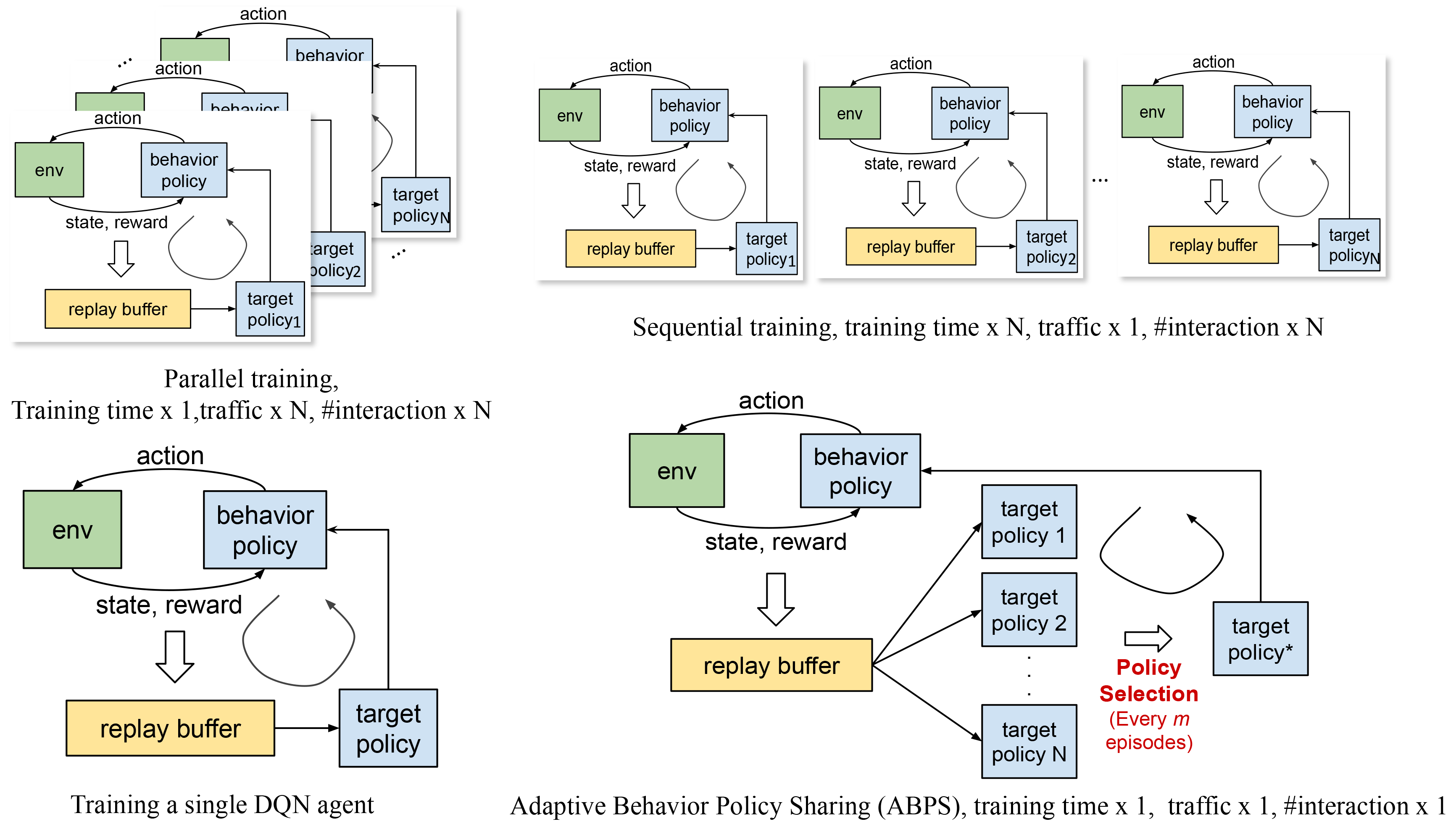}
\caption{Illustration of conventional hyper-parameter tuning with sequential or parallel independent training and hyper-parameter tuning with \methodname{}.}
\label{fig:diagram}
\end{figure}

Training the Q-network involves first enumerating a candidate set of hyper-parameters, where a pool of network architectures and/or optimization hyper-parameters such as learning rate, $\epsilon$ decay period, etc., are selected. The goal is to find the best set such that the agent trained with that hyper-parameter set achieves best online evaluation results. As shown in Figure~\ref{fig:diagram}, conventional hyper-parameter tuning requires $N$ times more interaction steps (where $N$ is the size of hyper-parameter pool) as each agent collects experience on its own and a separate replay buffer is used. Moreover, it often requires either $N$ times longer training time or more online traffic which is not desirable. 

There are several more practical constraints that need to be taken into consideration in real world production scale systems: 
\begin{enumerate}
\item The cost of deploying new model is high, and usually serving a huge ensemble of models is not desirable.
\item Frequent change of online model may cause inconsistent service to the user.
\end{enumerate}

\vskip -0.15in

\subsection{\methodname{}~(\methodabbrev{})}
With the above mentioned constraints, we propose \methodname{} (\methodabbrev{}) to improve the data-efficiency in hyper-parameter tuning by enabling sharing of experience among the agents while learning a \emph{policy selection strategy} $\sigma$ that selects behavior policy adaptively such that the value function networks are optimized efficiently. The overall work-flow of \methodabbrev{} is illustrated in Figure~\ref{fig:diagram}. A pool $\mathcal{P}$ of $K$ DQN action-value function networks $Q_1,\dots,Q_K$ are trained simultaneously with a \emph{shared} replay buffer $\mathcal{D}$, with each agent using a different hyper-parameter setup. During training, only \emph{one} of the agents will be deployed as behavior agent at each time step. An $\epsilon$-greedy policy of that agent is used to sample actions and the transitions are stored into the shared replay buffer using which all agents update their model parameters. After every $m$ episodes, the received return is used to evaluate the corresponding behavior policy so that $\sigma$ can be updated. Then a new behavior policy is chosen by $\sigma$ for the next $m$ episodes. This work-flow increases the stability of behavior policy during a period of $m$ episodes and reduces the frequency of policy selection compared to \citep{laroche2018}. However, it also increases the possibility of a bad behavior agent being chosen by chance which hurts the training of the rest by continuously collecting bad transitions. The effect of the policy selection period is studied in Section~\ref{result-big}.

Learning the strategy $\sigma$ is another exploration and exploitation trade-off problem. Inspired by \citep{bootstrap} where uniformly sampled Q-function from an ensemble of randomized Q-networks could effectively result in better exploration, we experiment with a fully random policy selection strategy $\sigma_{random}$ (naming it ABPS-random). Another natural selection for $\sigma$ is multi-armed bandit\cite{BubeckC12}, which has been used in many algorithm selection\cite{GaglioloS06,GaglioloS08} and meta-learning\cite{FialhoCSS10} problems. An UCB1 version of the bandit is often used out of the principle of optimism in the face of uncertainty. Here we study additional exploration strategies for the bandit problem and propose three variations of ABPS-bandit:
\begin{itemize}
    \item \textit{ABPS-bandit-UCB}, where the arm with largest UCB value is chosen.
    \item \textit{ABPS-bandit-$\epsilon$}, where with probability $\epsilon$ we randomly pick one agent from the pool and otherwise pick the arm with largest averaged reward.
    \item \textit{ABPS-bandit-softmax}, where the agents are sampled from soft-max probability of the averaged reward of each arm.
\end{itemize}
At the end of the training, an ensemble of agents is obtained. Among them, the one or more top-performed agents are chosen to be deployed for serving. Note that the training only takes the same amount of time and environmental interaction as a single agent training run, whereas a naive hyper-parameter tuning requires up to $K$ times more interactions and becomes less efficient. A full description of \methodabbrev{} algorithm can be found in Algorithm~\ref{abps}. To account for the non-stationarity in deep neural network training, we used a bandit with sliding window instead. Another alternative is using a discount rate for historical values.
\begin{algorithm}[hbt!]
\caption{ABPS-DQN training procedure}
\label{abps}
\begin{algorithmic}
\State Initialize DQN action-value function networks $Q_1$,$Q_2$,...,$Q_K$ 
\State Initialize shared replay buffer $\mathcal{D}$, initialize bandit $\mathcal{B}$ with $K$ arms
\For{t = 1,\dots, K}
    \State Run initial online evaluation for $Q_t$ for $n$ episodes, update arm $t$ of $\mathcal{B}$ with the averaged reward $\bar{X_t}=\bar{R_t}$
\EndFor
\Repeat
\If{ABPS\_TYPE = \texttt{random}} Randomly select one network as behavior $Q^*$ function
\ElsIf{ABPS\_TYPE = \texttt{soft-max}} Sample behavior $Q^*$ function with $p(Q_{i})=\frac{e^{\bar{X_i}}}{\sum_{j=1}^Ke^{\bar{X_j}}}$
\ElsIf{ABPS\_TYPE = \texttt{$\epsilon$-greedy}}
    \State With probability $\epsilon$ randomly select one network as behavior $Q^*$ function
    \State Otherwise select $Q_{I_t}$ as behavior $Q^*$ function where $I_t=\arg\max_{0\leq i\leq K}\bar{X_i}$
\ElsIf{ABPS\_TYPE = \texttt{UCB}}
    \State select $Q_{I_t}$ as behavior $Q^*$ function where $I_t=\arg \max_{0\leq i\leq K}\bar{X_i}+\sqrt{\xi\log{t}/{n_i}}$
\EndIf
\For{episode = 1,\dots,m}
    \Repeat
    \State Roll out $s$ steps with $\pi(Q^*)$ and store transitions in $\mathcal{D}$
    \State Train $Q_1$,\dots,$Q_K$ for 1 step
    \Until{episode ends}
\EndFor
\State Update arm $I_t$ of $\mathcal{B}$ with averaged $m$-episode reward $\bar{R}_{I_t}$ $$\bar{X_{I_t}}\leftarrow\frac{n_{I_t}\bar{X}_{I_t}+\bar{R}_{I_t}}{n_{I_t} + 1},\, n_{I_t}\leftarrow n_{I_t}+1$$
\State $t \leftarrow t+1$
\Until{iteration limit reached}
\end{algorithmic}
\end{algorithm}

\subsection{\methodname{} with evolving hyper-parameters}
Given the complexity of the hyper-parameter and neural architecture space, it is possible that the true global optimal is not covered in the empirically chosen pool of hyper-parameters. Moreover, the ideal hyper-parameters for deep learning problems are themselves highly non-stationary. Recent studies have proposed many techniques to conquer such deficiency, one of the successful approaches of such is called Population Based Training (PBT)\citep{jaderberg2017population,Li2019}. The idea is to use information from the rest of the population to refine the hyper-parameters through periodically substituting bad workers with promising workers (exploit) and random evolution of hyper-parameters (explore). However, each worker in the pool is trained independently, which again requires $K$ times interactions with the environment where $K$ is the number of workers. Given that both PBT and \methodabbrev{} utilizes sharing of information across the population and deals with an exploitation/exploration problem, we propose a hybrid of the two, named \methodpbt{}, such that both data efficiency and hyper-parameter search efficiency can be achieved. 

\begin{figure}[hbt!]
\centering
\vskip -0.1in
  \includegraphics[width=0.75\linewidth,trim=0cm 0cm 0cm 0cm, clip=true]{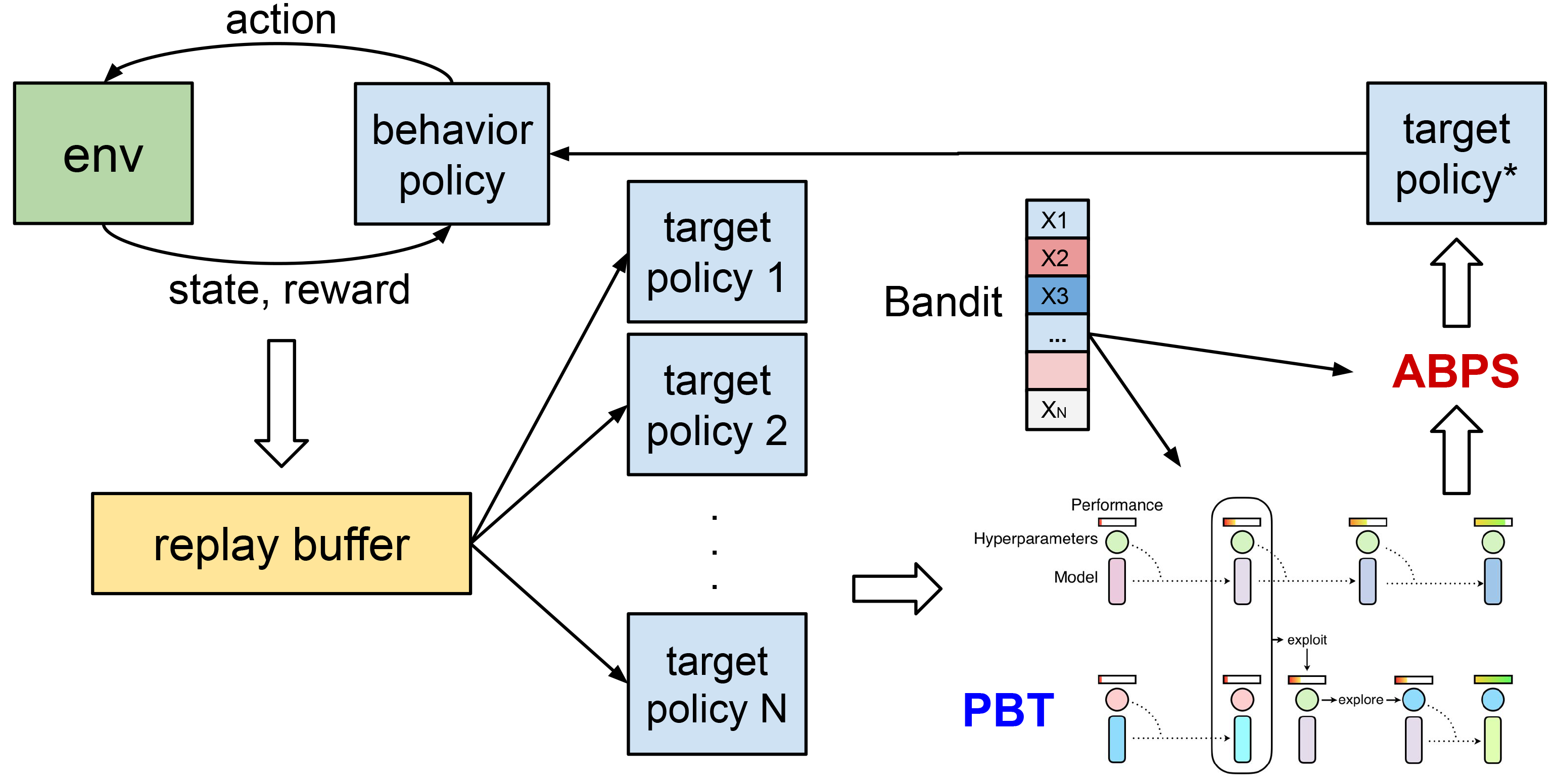}
\caption{Illustration of \methodabbrev{} combined with Population Based Training}
\label{fig:abps-pbt}
\vskip -0.1in
\end{figure}

Figure~\ref{fig:abps-pbt} shows the overall work-flow of \methodpbt{}. Some adaptations are required to make PBT compatible with ABPS. The vanilla PBT exploitation uses real-time online evaluation for all workers in the pool, while in \methodpbt{} the evaluation is replaced by the ABPS-bandit value tables. Even though the sliding window bandit keeps relatively up-to-date values for each arm, we set an additional threshold on the last update time to detect outdated values. An online evaluation will be triggered immediately when the arm is not sufficiently up-to-date for PBT to do useful exploitation. Both PBT and ABPS has a period set between the triggering of algorithms. We set the period for PBT to be a multiples of the period for ABPS, since ABPS's policy selection happens more frequently than PBT-exploitation. In addition to model parameters and hyper-parameters, the bad worker's bandit value is also substituted by the counterparts of the good worker during PBT-exploitation. A detailed description of the algorithm can be found in Algorithm~\ref{abps-pbt}.

\begin{algorithm}[hbt!]
\caption{ABPS-PBT-DQN training procedure}
\label{abps-pbt}
\begin{algorithmic}
\State Initialize population pool $\mathcal{P}$ of $K$ DQN action-value function networks $Q_1$,\dots ,$Q_K$ associated with $K$ sets of hyper-parameters $h_1$,$h_2$,...,$h_K$
\State Initialize shared replay buffer $\mathcal{D}$, initialize bandit $\mathcal{B}$ with $K$ arms
\State Run initial evaluations and update $\mathcal{B}$ 
\Repeat
\State Select behavior policy $\pi(Q^*)$ with ABPS (described in Algorithm 1)
\For{episode = 1,\dots,m}
    \Repeat
    \State Roll out $s$ steps with $\pi(Q^*)$ and store transitions in $\mathcal{D}$
    \State Train $Q_1$,\dots,$Q_K$ for 1 step
    \Until{episode ends}
\EndFor
\State Update the mean value of the chosen arm with averaged episode reward (see Algorithm 1)
\For{each agent $Q_i$}
    \If{\texttt{pbt-ready}} 
        \If{arm $i$ not recently updated}   run $\texttt{eval}(Q_i)$ and update $\bar{X_i}$ \EndIf
        \State Exploit using the bandit values: $h_i',\theta_i',\bar{X_i}' \leftarrow \texttt{exploit}(h_i,\theta_i,\bar{X_i}, \mathcal{P})$
        \If{$\theta_i\neq\theta_i'$}    $h_i,\theta_i\leftarrow \texttt{explore}(h_i',\theta_i',\mathcal{P})$ \EndIf
    \EndIf
\EndFor
\State $t \leftarrow t+1$
\Until{iteration limit reached}
\end{algorithmic}
\end{algorithm}

\subsection{Algorithm evaluation}
Instead of examining the behavior policy reward collected during training (\textit{i.e.}, the regret of the behavior policy), we run a separate online evaluation for $50$ episodes for every agent at each training epoch. The online evaluation reward reflects the goodness of the individual agent in the pool. We choose the agent with the best online evaluation reward at the end of training, and compare it with the best agent achieved from full hyper-parameter tuning without \methodabbrev{}. This simulates the real world scenario where we care about not only the regret during training but also the serving performance. To evaluate the overall performance of the ensemble of the agents, we also look at the top $25\%$ quantile and the variance of the rewards.

\section{Experiments}
\label{result}

\subsection{DQN with a small hyper-parameter pool}
\label{result-small}
\begin{figure}[hbt!]
\centering
  \includegraphics[width=0.9\linewidth,trim=0cm 0cm 0cm 0cm, clip=true]{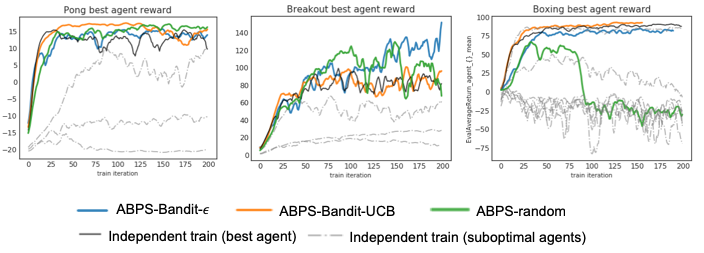}
\caption{Best agent online evaluation performance on Atari Pong (left), Breakout (middle) and Boxing (right) for hyper-parameter pool of size 4 (left, middle) and 8 (right, 6 out of 8 are small networks). Black line represents the best agent achieved by traditional hyper-parameter tuning without \methodabbrev{} and dashed grey lines are the sub-optimal agents in the hyper-parameter pool trained without \methodabbrev{}.}
\label{fig:small-results}
\end{figure}

We first experiment with a small hyper-parameter pool with 4 candidate network architectures:
\begin{itemize}
    \item Agent normal(\cite{mnih2013playing}): (16, (8, 8)) -> (32, (4, 4)) -> 256
    \item Agent small: (8, (8, 8)) -> (8, (4, 4)) -> 32
    \item Agent wide: (32, (5, 5)) -> (64, (3, 3)) -> 1024
    \item Agent deep: (32,(8, 8)) -> (64, (4, 4)) -> (64, (3, 3)) -> 512
\end{itemize}
Where numbers in parenthesis represent the number of convolutional filters and their corresponding shapes, and numbers on the most right represent the size of fully connected layer. We trained an ensemble of 4 agents with each using one of the candidate architectures on Atari Pong and Breakout, and an ensemble of 8 agents with 6 of them using variation of small architectures on Boxing to test the effect of bad agents on the different ABPS algorithms. Figure~\ref{fig:small-results} shows the best agent performance of 3 ABPS variations and the performance of all agents trained independently without ABPS. ABPS-bandit methods achieved better evaluation performance on all three games with almost equivalent convergence speed with the independent training of the best agent in the pool. ABPS-Bandit-UCB converges better and faster than ABPS-Bandit-$\epsilon$ on Pong and Boxing. Surprisingly, even a random policy selection strategy could result in the same level of performance as ABPS-Bandit, demonstrating the power of exploration through ensemble bootstrapping effect. However, when the hyper-parameter pool is dominated with sub-optimal choices, the random algorithm is largely affected by bad agents, resulting in a drastic drop of best agent performance.

\subsection{DQN with a large hyper-parameter pool}
\label{result-big}

\begin{figure}[hbt!]
\centering
  \includegraphics[width=1\linewidth,trim=0cm 0cm 0cm 0cm, clip=true]{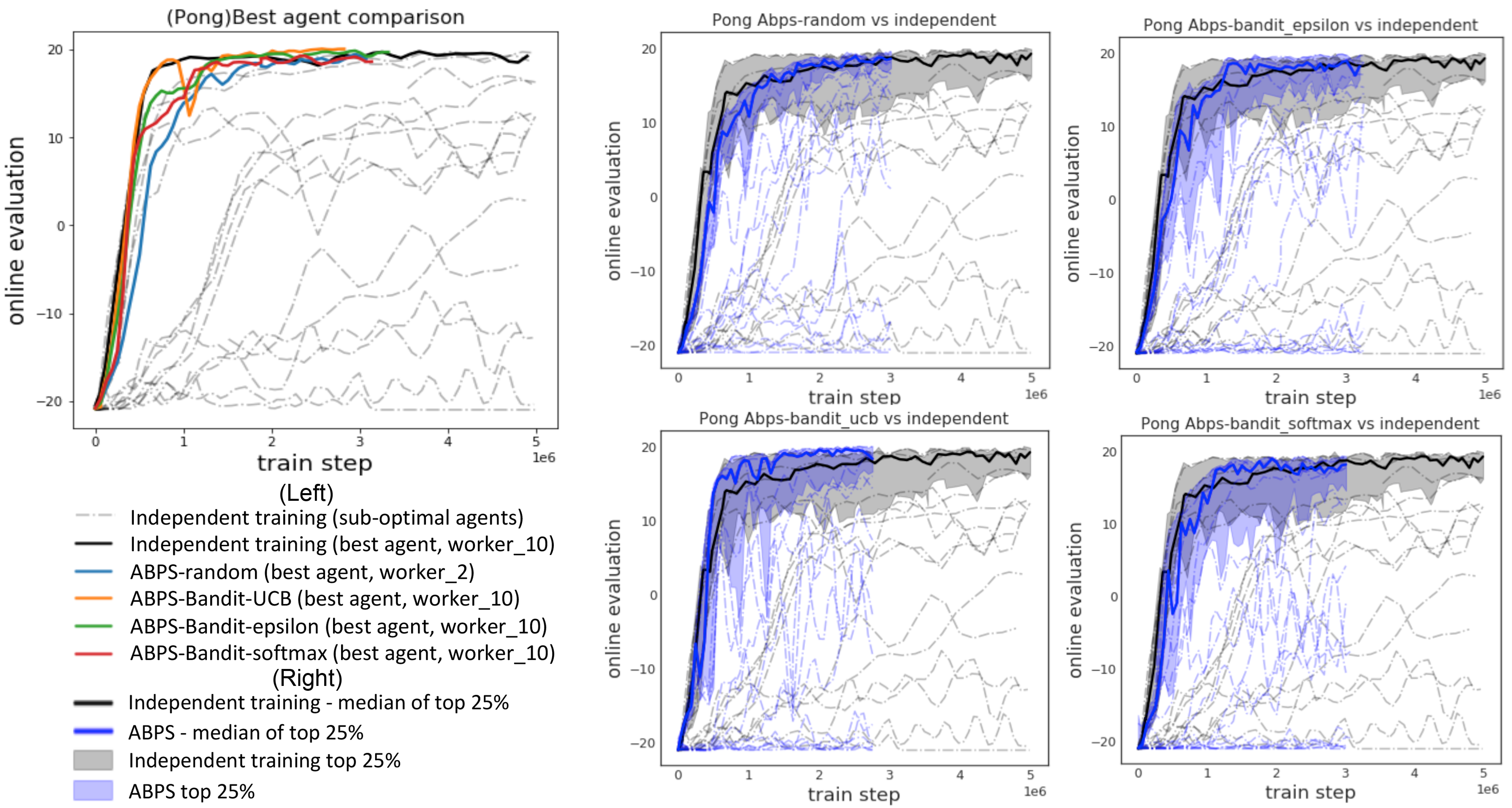}
\caption{Best agent online evaluation (left) and top $25\%$ quantile (right) of the online evaluation performance of the whole ensemble of hyper-parameter pools. A pool of $16$ agents was used with each of them using hyper-parameters sampled from a prior distribution. Black line represents the best agent achieved by traditional hyper-parameter tuning without \methodabbrev{} and dashed grey lines are the sub-optimal agents in the hyper-parameter pool trained without \methodabbrev{}.}
\label{fig:big-result}
\vskip -0.0in
\end{figure}

We further evaluate the performance of ABPS on large hyper-parameter pool with various architectures and optimization hyper-parameters on Atari Pong. A pool of $16$ agents is drawn from a prior distribution, where the marginal probability over the four architectures is $[0.2,0.2,0.2,0.4]$ for normal, wide, deep and small correspondingly, and a random perturbation (scaled by $0.9\sim1.1$) is made to the number of convolutional and fully connected units. The learning rate $\lambda$ and $\epsilon$ decay period is log-uniformly sampled from $[0.00001, 0.005]$ and $[2.5e5, 4e6]$ respectively. Figure~\ref{fig:big-result} shows the online evaluation performances of best agent and top $25\%$ agents through out the training, where all ABPS variants achieved equivalently good convergence as the best hyper-parameter in the pool and ABPS-UCB achieved the fastest convergence speed. The best agent selected by ABPS-bandit methods is the same as the best in independent hyper-parameter tuning, showing ABPS's capability to recover the traditional hyper-parameter tuning results (if not better) with almost no computation overhead and much fewer environmental interactions. We also discovered ABPS's capability of improving the overall performance of the ensemble, with higher top $25\%$ quantile and lower variance.
\vskip -0.1in

\begin{figure}[hbt!]
\centering
  \includegraphics[width=1\linewidth,trim=0cm 0cm 0cm 0cm, clip=true]{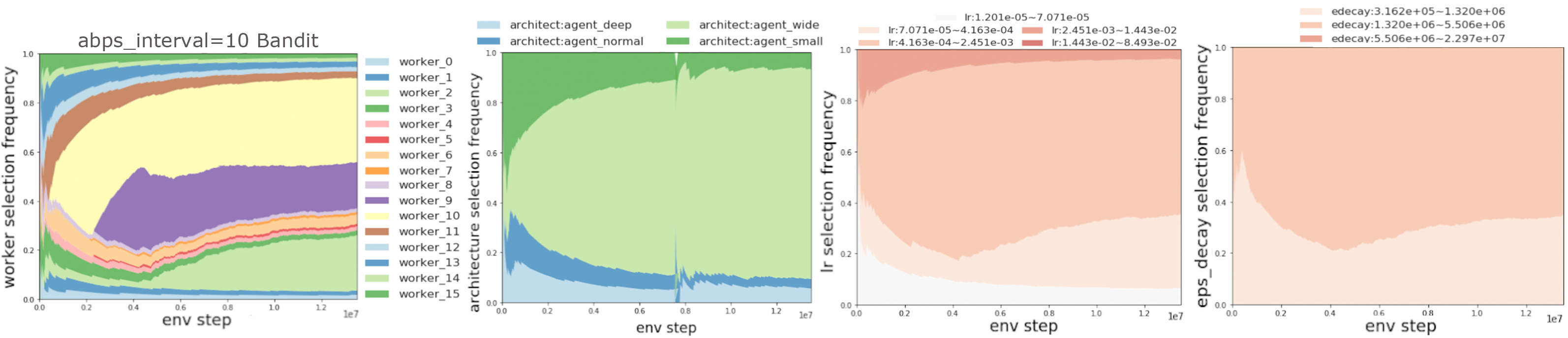}
\caption{Frequency of different workers (top), architecture (left), learning rate (middle) and epsilon decay speed (right) being chosen as behavior policy at each time step.}
\label{fig:big-hyper}
\end{figure}

Figure~\ref{fig:big-hyper} shows the accumulated frequency of agents with different architectures and optimization hyper-parameters being chosen as the behavior policy from ABPS-Bandit-softmax. We observed an increasing frequency of using wider networks against the others, and the adaptation of hyper-parameters. Larger learning rate and high $\epsilon$ are preferred on the earlier training rounds and are both reduced at later training rounds, which is consistent with the annealing strategy in prior studies on the optimization of deep RL.
\begin{figure}[hbt!]
\centering
  \includegraphics[width=1.0\linewidth,trim=1cm 0cm 0.3cm 0cm, clip=true]{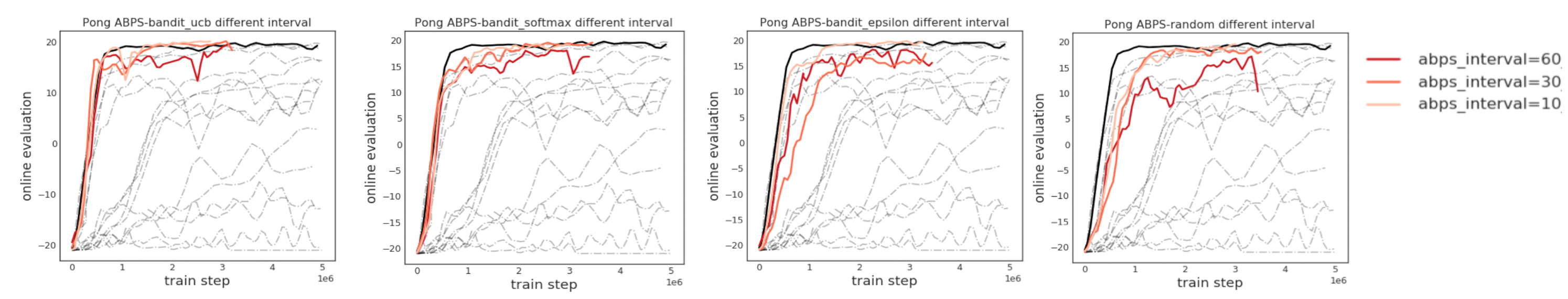}
\caption{Effect of policy selection period length on the ABPS training using different exploration strategies (UCB, softmax, $\epsilon$-greedy and random from left to right). ABPS variations that rely on value based exploration (UCB, softmax) are less effected than strategies that rely on random exploration (random, $\epsilon$-greedy)}
\label{fig:big-period}
\end{figure}

We also study the effect of policy selection period on ABPS performance and the result is shown in Figure~\ref{fig:big-period}. We found that ABPS variations that rely on random exploration (random, $\epsilon$-greedy) are more affected by longer selection period than value based exploration (UCB, softmax). With up to $60$ episodes of ABPS interval, the ABPS-Bandit-UCB still achieves comparable performance with traditional hyper-parameter tuning.

\subsection{Integrating ABPS with Population Based Training}
\begin{figure}[hbt!]
\centering
  \includegraphics[width=1.0\linewidth,trim=0cm 0cm 0cm 0cm, clip=true]{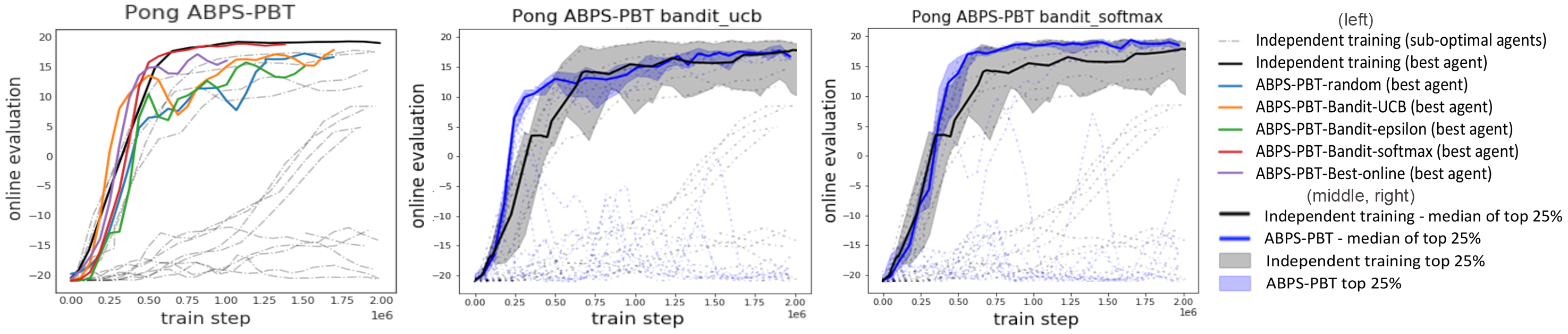}
\caption{Best agent online evaluation (left) and top $25\%$ quantile (right) of the online evaluation performance of the whole ensemble of hyper-parameter pools using ABPS-PBT.}
\label{fig:big-pbt}
\vskip -0.1in
\end{figure}

With the same experimental setup as section~\ref{result-big}, we evaluate the performance of ABPS-PBT. As shown by Figure~\ref{fig:big-pbt}, the best agent convergence speed is dramatically improved after combining with PBT, and the top $25\%$ agents are significantly improved with much smaller variance and higher median, showing the effectiveness of ABPS-PBT.

\section{Conclusion}
We propose Adaptive Behavior Policy Sharing, a data-efficient training algorithm that allows sharing of experience collected by adaptively selected behavior policies. We further adapt Population Based Training and proved that it is compatible with ABPS. We experimented with multiple Atari games with up to $16$ hyper-parameter/architecture setups, and achieved superior overall performance, smaller variance on top $25\%$ agents, and matching performance on the best agent compared to conventional hyper-parameter tuning with independent training, even though \methodabbrev{} only needs the same number of environmental interactions as training a single agent. We also show that \methodpbt{} has the potential to improve both the convergence speed and the variance. 

\bibliography{abps.bib}
\bibliographystyle{plainnat}
\clearpage


\end{document}